\documentclass[letterpaper, 10 pt, conference]{ieeeconf}  % Comment this line out if you need a4paper

\IEEEoverridecommandlockouts                              % This command is only needed if 
\usepackage{graphics} 
\usepackage{epsfig}
\usepackage{mathptmx} 
\usepackage{times} 

\usepackage[ruled,vlined,linesnumbered]{algorithm2e}
\usepackage{amsmath,amssymb,amsfonts}
\usepackage{subcaption}
\usepackage{algorithmic}
\usepackage{color}
\usepackage[font={small},belowskip=0pt,aboveskip=3pt]{caption}
\usepackage{comment}

\newcommand{\T}{^{\mbox{\tiny\sf T}}}

\newcommand{\m}{^{\mbox{\scriptsize{-}}}}   

\newtheorem{theorem}{Theorem}

\newtheorem{proposition}[theorem]{Proposition}

\title{\LARGE \bf
IBBT: Informed Batch Belief Trees for Motion Planning \\
Under Uncertainty
}

\author{Dongliang Zheng$^{1}$ and Panagiotis Tsiotras$^{2}$% <-this % stops a space
\thanks{This work has been supported by NSF award IIS-2008695 and by Ford through the Georgia Tech/Ford Alliance program.}% <-this % stops a space
\thanks{$^{1}$Dongliang Zheng is a Ph.D. student with the School of Aerospace Engineering,
        Georgia Institute of Technology, Atlanta, GA 30332, USA. Email:
        {\tt\small dzheng@gatech.edu}}%
\thanks{$^{2}$Panagiotis Tsiotras is a Professor with the School of Aerospace Engineering and Institute for Robotics and Intelligent Machines, Georgia Institute of Technology, Atlanta, GA 30332, USA. Email:
        {\tt\small tsiotras@gatech.edu}}%
}

\begin{document}

\maketitle
\thispagestyle{empty}
\pagestyle{empty}

%%%%%%%%%%%%%%%%%%%%%%%%%%%%%%%%%%%%%%%%%%%%%%%%%%%%%%%%%%%%%%%%%%%%%%%%%%%%%%%%
\begin{abstract}

In this work, we propose the Informed Batch Belief Trees (IBBT) algorithm for motion planning under motion and sensing uncertainties.
The original stochastic motion planning problem is divided into a \textit{deterministic} motion planning problem and a graph search problem. We solve the deterministic planning problem using sampling-based methods such as PRM or RRG to construct a graph of nominal trajectories. Then, an informed cost-to-go heuristic for the original problem is computed based on the nominal trajectory graph. 
Finally, we grow a belief tree by searching over the graph using the proposed heuristic.    
IBBT interleaves between batch state sampling, nominal trajectory graph construction, heuristic computing, and search over the graph to find belief space motion plans.  
IBBT is an anytime, incremental algorithm. 
With an increasing number of batches of samples added to the graph, the algorithm finds motion plans that converge to the optimal one. IBBT is efficient by reusing results between sequential iterations. The belief tree searching is an ordered search guided by an informed heuristic.
We test IBBT in different planning environments. 
Our numerical investigation confirms that IBBT finds non-trivial motion plans and is faster compared with previous similar methods.
\end{abstract}
%\begin{IEEEkeywords}
%Motion planning, belief space planning, Planning under uncertainty
%\end{IEEEkeywords}

%%%%%%%%%%%%%%%%%%%%%%%%%%%%%%%%%%%%%%%%%%%%%%%%%%%%%%%%%%%%%%%%%%%%%%%%%%%%%%%%
\section{Introduction}

For safe and reliable autonomous robot operation in a real-world environment, consideration of various uncertainties becomes necessary.
These uncertainties may arise from an inaccurate motion model, actuation or sensor noise, partial sensing, and the
presence of other agents moving in the same environment.  
In this paper, we study the safe motion planning problem for robot systems with nontrivial dynamics, motion uncertainty, and state-dependent measurement uncertainty in an environment with non-convex obstacles. 

Planning under uncertainty is referred to as belief space planning (BSP), where the state of the robot is characterized by a probability distribution function (pdf) over all possible states.
This pdf is commonly referred to as the \textit{belief} or
information state~\cite{Thrun2005Probabilistic,Van2012Motion}.
A BSP problem can be formulated as a partially observable Markov decision process (POMDP) problem \cite{Kaelbling1998Planning}.
Solving POMDPs for continuous state, control, and observation spaces, is, however, intractable.
Existing methods based on discretization are resolution-limited~\cite{Porta2006Point, shani2013survey}. 
Optimization over the entire discretized belief space to find a path is computationally expensive and does not scale well to large-scale problems. 
Online POMDP algorithms are often limited to short-horizon planning, have challenges when dealing with local minima, and are not suitable for global planning in large environments~\cite{somani2013despot} % ~\cite{ross2008online, somani2013despot}. 

Planning in infinite-dimensional distributional (e.g., belief) spaces can become more tractable by using sampling-based methods~\cite{Karaman2011Sampling}.
For example, belief roadmap methods~\cite{Prentice2009The} build a belief roadmap to reduce estimation uncertainty;
the rapidly-exploring random belief trees (RRBT) algorithm~\cite{Bry2011Rapidly} has been  proposed to grow a tree in the belief space.
Owing to their advantages in avoiding local minima, dealing with nonconvex obstacles and high-dimensional state spaces, along with their anytime property, sampling-based methods have gained increased attention in the robotics community~\cite{Luders2013Robust, Sun2015High, Janson2018Monte, Ichter2017Real, Agha-Mohammadi2014FIRM}.

Robot safety under uncertainty can be also formulated as a chance-constrained optimization problem~\cite{Blackmore2011Chance, Vitus2011Closed, Wang2020Non-Gaussian, Bry2011Rapidly}.
In addition to minimizing the cost function, one also wants the robot not to collide with obstacles, with high probability. 
By approximating the chance constraints as deterministic constraints,  references \cite{Blackmore2011Chance, Vitus2011Closed, Wang2020Non-Gaussian} solve the problem using an optimization-based framework.
However, those approaches lack scalability with respect to problem complexity~\cite{Aoude2013Prob}, and the explicit representation of the obstacles is usually required.

In this paper, we focus on sampling-based approaches similar to~\cite{Bry2011Rapidly, Luders2013Robust, Summers2018Distributionally}.
One challenge of sampling-based algorithms for planning under uncertainty is the lack of the optimal substructure property, which has been discussed in~\cite{Bry2011Rapidly, Agha-Mohammadi2014FIRM,Zheng2021Belief}.
The lack of optimal substructure property is further explained by the lack of total ordering on paths based on cost.
Specifically, it is not enough to only minimize the usual cost function ~--~ explicitly finding paths that reduce the uncertainty of the robot is also important (see Figure~\ref{BBTmotivation}(a)).
The RRBT algorithm proposed in \cite{Bry2011Rapidly} overcomes the lack of optimal substructure property by introducing a partial-ordering of belief nodes and by keeping all non-dominated nodes in the belief tree.
Note that without this partial-ordering, the methods in~\cite{Luders2013Robust, Sun2015High, Janson2018Monte, Summers2018Distributionally} may not be able to find a solution, even if one exists.
Minimizing the cost and checking the chance constraints can only guarantee that the existing paths in the tree satisfy the chance constraints.
Without searching for paths that explicitly reduce state uncertainty, it will be difficult for future paths to satisfy the chance constraints.

In this paper, we propose the Informed Batch Belief Tree (IBBT) algorithm, which improves over the RRBT algorithm with the introduction of \textit{batch sampling} and \textit{ordered graph search guided by an informed heuristic}.
Firstly, IBBT uses the partial ordering of belief nodes as in~\cite{Bry2011Rapidly}.
Compared to~\cite{Luders2013Robust, Sun2015High, Janson2018Monte, Summers2018Distributionally}, IBBT is able to find sophisticated plans that visit and revisit the information-rich region to gain information.
Secondly, RRBT uses unordered search like RRT* while IBBT uses batch sampling and ordered search. RRBT adds one sample each time to the graph randomly.
As shown in \cite{janson2015fast} and  \cite{gammell2020batch}, ordered searches such as FMT* and BIT* perform better than RRT*.
Thirdly, RRBT only uses the cost-to-come cost to guide the belief tree search while IBBT introduces a cost-to-go heuristic and uses the total path cost heuristic for informed belief tree search.
After adding a sample, RRBT performs an exhaustive graph search. Thus all non-dominated belief nodes are added to the belief tree.   
With batch sampling and informed graph search, IBBT avoids adding unnecessary belief nodes. 
Thus, IBBT is able to find the initial solution in a shorter time and has better cost-time performance compared to RRBT.% , as will be shown in Section~\ref{SecExperiment}.

% The remainder of the paper is organized as follows. In Section~\ref{SecRelatedWorks}, some related works are discussed.
% The problem formulation is given in Section~\ref{SecProblemformulation}. 
% Some preliminary results are presented in Section~\ref{Sec:Preliminary}.
% The Batch Belief Tree algorithm is given in Section~\ref{SecBBT}. 
% The experimental results of BBT along with the comparison with other methods is given in Section~\ref{SecExperiment}.
% Finally, Section~\ref{secConclusion} concludes the paper.

\section{Related Works} \label{SecRelatedWorks}

In~\cite{Prentice2009The}, the problem of finding the minimum estimation uncertainty path for a robot from a starting position to a goal is studied by building a roadmap. 
In~\cite{Bry2011Rapidly, Van2011LQG-MP}, it was noted that the true \textit{a priori} probability distribution of the state should be used for motion planning instead of assuming maximum likelihood observations~\cite{Prentice2009The,Platt2010Belief}.
A linear-quadratic Gaussian (LQG) controller along with the RRT algorithm~\cite{lavalle2001randomized} were used for motion planning in \cite{Van2011LQG-MP}.
To achieve asymptotic optimality, the authors in \cite{Bry2011Rapidly} incrementally construct a graph and search over the graph to find all non-dominated belief nodes.
Given the current graph, the Pareto frontier of belief nodes at each vertex is saved, where the Pareto frontier is defined by considering both the path cost and the node uncertainty.

In~\cite{Sun2015High} high-frequency replanning is shown to be able to better react to uncertainty during plan execution.
Monte Carlo simulation and importance sampling are used in~\cite{Janson2018Monte} to compute the collision probability.
Moving obstacles are considered in~\cite{Aoude2013Prob}.
In~\cite{Liu2014Incremental}, state dependence of the collision probability is considered and incorporated with chance-constraint RRT*~\cite{Luders2013Robust, Luders2010Chance}. 
In~\cite{Shan2017Belief}, a roadmap search method is proposed to deal with localization uncertainty; 
however, solutions for which the robot needs to revisit a position to gain information are ruled out.
Distributionally robust RRT is proposed in~\cite{Summers2018Distributionally, Safaoui2021Risk}, where moment-based ambiguity sets of distributions are used to enforce chance constraints instead of assuming Gaussian distributions.
Similarly, a moment-based approach that considers non-Gaussian state distributions is studied in~\cite{Wang2020Moment}. In \cite{Ho2022Gaussian}, the Wasserstein distance % \cite{Panaretos2019Statistical} 
is used as a metric for Gaussian belief space planning. The algorithm is compared with RRBT. However, from the simulation results, RRBT usually finds better (lower cost) plans and thus has a better convergence performance. 

Other works that are not based on sampling-based methods formulate the chance-constrained motion planning problem as an optimization problem~\cite{Blackmore2011Chance, Vitus2011Closed, Wang2020Non-Gaussian}.
In those methods, the explicit representation of the obstacles is usually required. 
The obstacles may be represented by convex constraints or polynomial constraints.
The chance constraints are then approximated as deterministic constraints and the optimization problem is solved by convex~\cite{Vitus2011Closed} or nonlinear programming~\cite{Wang2020Non-Gaussian}. 
Differential dynamic programming has also been used to solve motion planning under uncertainty~\cite{Van2012Motion, Sun2021Belief, Rahman2021Uncertainty}.
These algorithms find a locally optimal trajectory in the neighborhood of a given reference trajectory.
The algorithms iteratively linearize the system dynamics along the reference trajectory and solve an LQG problem to find the next reference trajectory.

\section{Problem formulation} \label{SecProblemformulation}

We consider the problem of planning for a robot with nontrivial dynamics, model uncertainty, measurement uncertainty from sensor noise, and obstacle constraints.
The state-space $\mathcal{X}$ is decomposed into free space $\mathcal{X}_\mathrm{free}$ and obstacle space $\mathcal{X}_\mathrm{obs}$.
The motion planning problem is given by 
\begin{align}
& \mathop{\arg\min}_{u_k} \ \mathbb{E}\left[\sum_{k=0}^{N-1} J(x_k, u_k) \right], \label{eq:obj}\\
\mathrm{s.t.} \ \ & x_0 \sim \mathcal{N}(\bar{x}_s,P_0), \ \bar{x}_N=\bar{x}_g, \label{eq:constraint1}\\
& P(x_k \in \mathcal{X}_\mathrm{obs}) < \delta, \ k = 0, \cdots, N, \label{eq:constraint2}\\
& x_{k+1} = f(x_k,u_k,w_k), \ k=0,\ldots,N-1, \label{eq:nonlinearModel}\\
& y_k = h(x_k,v_k), \ k=0,\ldots,N-1, \label{eq:nonlinearModel1}
\end{align}
where (\ref{eq:nonlinearModel}) and (\ref{eq:nonlinearModel1}) are the motion and sensing models, respectively. 
Furthermore, $x_k \in \mathbb{R}^{n_x}$ is the state, $u_k \in \mathbb{R}^{n_u}$ is the control input, and $y_k \in \mathbb{R}^{n_y}$ is the measurement at time step $k = 0,1,\ldots,N-1$,
where the steps of the noise processes $w_k \in \mathbb{R}^{n_w}$ and $v_k \in \mathbb{R}^{n_y}$ are i.i.d standard Gaussian random vectors, respectively.
We assume that $(w_k)_{k=0}^{N-1}$ and $(v_k)_{k=0}^{N-1}$ are independent.
Expression (\ref{eq:constraint1}) is the boundary condition for the motion planning problem. 
The goal is to steer the system from some initial distribution to a goal state. 
Since the robot state is uncertain, the mean of the final state $\bar{x}_N$ is constrained to be equal to the goal state $\bar{x}_g$.
Condition (\ref{eq:constraint2}) is a chance constraint that enforces safety of the robot.

Similar to \cite{Bry2011Rapidly}, the motion plan considered in this paper is formed by a nominal trajectory and a feedback controller that stabilizes the system around the nominal trajectory.  
Specifically, we will use a \texttt{Connect} function that returns a nominal trajectory and a stabilizing controller between two states $\bar{x}^a$ and $\bar{x}^b$,
\begin{equation}
		(\bar{X}^{a,b}, \bar{U}^{a,b}, K^{a,b}) = \texttt{Connect}(\bar{x}^a, \bar{x}^b),
\end{equation}
$\bar{X}^{a,b}$ and $\bar{U}^{a,b}$ are the sequence of states and controls of the nominal trajectory, and
$K^{a,b}$ is a sequence of the corresponding feedback control gains.
The nominal trajectory can be obtained by solving a deterministic optimal control problem with boundary conditions $\bar{x}^a$ and $\bar{x}^b$, and system dynamics $\bar{x}_{k+1} = f(\bar{x}_k,\bar{u}_k,0)$.
The stabilizing controller can be computed using, for example, finite-time LQR design~\cite{Agha-Mohammadi2014FIRM}.

A Kalman filter is used for online state estimation, which gives the state 
estimate\footnote{Note non-standard notation.} $\hat{x}_k$ of $(x_k - \bar{x}_k)$.
Thus, the control at time $k$ is given by
\begin{align}
		u_k = \bar{u}_k + K_k \hat{x}_k. \label{eq:feedbackControl}
\end{align}

With the introduction of the \texttt{Connect} function, the optimal motion planning problem (\ref{eq:obj})-(\ref{eq:nonlinearModel1}) is reformulated as finding the sequence of intermediate states $(\bar{x}^0, \bar{x}^1, \cdots, \bar{x}^{\ell})$.
The final control is given by
\begin{align}
		(u_k)_{k=0}^{N-1} = (\texttt{Connect}(\bar{x}^0, \bar{x}^1), \cdots, \texttt{Connect}(\bar{x}^{\ell - 1}, \bar{x}^{\ell})).
\end{align}
The remaining problem is to find the optimal sequence of intermediate states and enforce the chance constraints (\ref{eq:constraint2}).

\section{Covariance Propagation} \label{Sec:Preliminary}

We assume that the system given by (\ref{eq:nonlinearModel}) and (\ref{eq:nonlinearModel1}) is locally well approximated by its linearization along the nominal trajectory. 
This is a common assumption as the system will stay close to the nominal trajectory using the feedback controller \cite{Agha-Mohammadi2014FIRM, Zheng2021Belief}.
Define 
\begin{subequations}
\begin{align}
    \check{x}_k &= x_k - \bar{x}_k, \label{a}\\
    \check{u}_k &= u_k - \bar{u}_k, \label{b}\\
    \check{y}_k &= y_k - h(\bar{x}_k, 0). \label{c}
\end{align}
\end{subequations}
By linearizing along $(\bar{x}_k,\bar{u}_k)$, the error dynamics is
\begin{equation}
\begin{split}
    \check{x}_k &= A_{k-1} \check{x}_{k-1} + B_{k-1} \check{u}_{k-1} + G_{k-1} w_{k-1}, \\
    \check{y}_k &= C_k \check{x}_k + D_k v_k.
    \label{EQ:LTV}
\end{split}
\end{equation}
We will consider this linear time-varying system hereafter.
A Kalman filter is used for estimating $\check{x}_k$ and is given by
\begin{align}
    \hat{x}_{k} & = \hat{x}_{k\m} + L_k (\check{y}_k - C_k \hat{x}_{k\m}), \label{KFdynamics}\\
    \hat{x}_{k\m} & = A_{k-1} \hat{x}_{k-1} + B_{k-1} \check{u}_{k-1},
    \label{KFdynamics1}
\end{align}
where, 
\begin{subequations}
\begin{align}
L_k & =\tilde{P}_{k\m} C_k \T ( C_k \tilde{P}_{k\m} C_k \T + D_k D_k \T )^{-1}, \label{KFupdatea} \\
\tilde{P}_k & =( I - L_k C_k) \tilde{P}_{k\m}, \label{KFupdateb}\\
\tilde{P}_{k\m} & = A_{k-1} \tilde{P}_{k-1} A_{k-1} \T +G_{k-1} G_{k-1} \T,
\label{KFupdatec}
\end{align}
\end{subequations}
and $L_k$ is the Kalman gain. 
The covariances of $\check{x}_k$, $\hat{x}_k$ and $\tilde{x}_k \triangleq \check{x}_k - \hat{x}_k $ are denoted as $P_k = \mathbb{E}[\check{x}_k \check{x}_k \T]$, $\hat{P}_k = \mathbb{E}[\hat{x}_k \hat{x}_k \T]$ and $\tilde{P}_k = \mathbb{E}[\tilde{x}_k \tilde{x}_k \T]$, respectively. 
Note that the covariance of $x_k$ is also given by $P_k$ and the estimation error covariance $\tilde{P}_k$ is computed from (\ref{KFupdateb}). 
From (\ref{EQ:LTV})-(\ref{KFdynamics1}), it can be verified that $\mathbb{E}[\check{x}_k] = \mathbb{E}[\hat{x}_k] = \mathbb{E}[\hat{x}_{k\m}]$. 
Since $\mathbb{E}[\check{x}_0] = 0$, by choosing $\mathbb{E}[\hat{x}_0] = 0$, we have $\mathbb{E}[\hat{x}_k] = 0$ for $k = 0, \cdots, N$.
Using (\ref{KFdynamics}) and (\ref{KFdynamics1}) we also have that
\begin{equation}
\begin{split}
    \hat{P}_k &= \mathbb{E}[\hat{x}_k \hat{x}_k \T]  \\
    &= \mathbb{E}[\hat{x}_{k\m} \hat{x}_{k\m} \T] + L_k (C_k \tilde{P}_{k\m} C_k \T + D_k D_k \T) L_k\T  \label{estimatedstateCov}\\
    &= (A_{k-1}+B_{k-1}K_{k-1}) \hat{P}_{k-1} (A_{k-1}+B_{k-1}K_{k-1}) \T + L_k C_k \tilde{P}_{k\m}
\end{split}
\end{equation}
Using the fact that $\mathbb{E}[\hat{x}_k \tilde{x}_k \T] = 0$, it can be verified that $P_k = \hat{P}_k + \tilde{P}_k$.
Thus, given the feedback gains $K_k$ and the Kalman filter gain $L_k$, we can predict the covariances of the state estimation error and the state along the trajectory, which also provides the state distributions in the case of a Gaussian distribution.

\section{Informed Batch Belief Tree Algorithm} \label{SecBBT}
\subsection{Motivation}
\begin{figure}[htb]
    \centering
    \begin{subfigure}[b]{0.49\columnwidth}
         \centering
         \includegraphics[width=1\columnwidth]{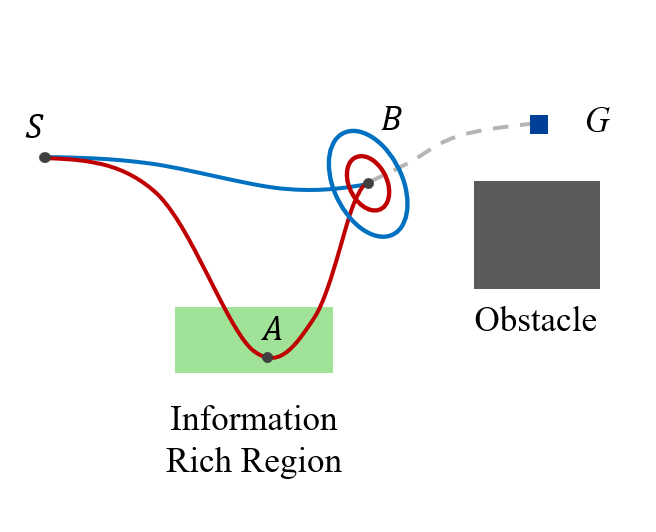}
         \caption{}
     \end{subfigure}
     \begin{subfigure}[b]{0.49\columnwidth}
         \centering
         \includegraphics[width=1\columnwidth]{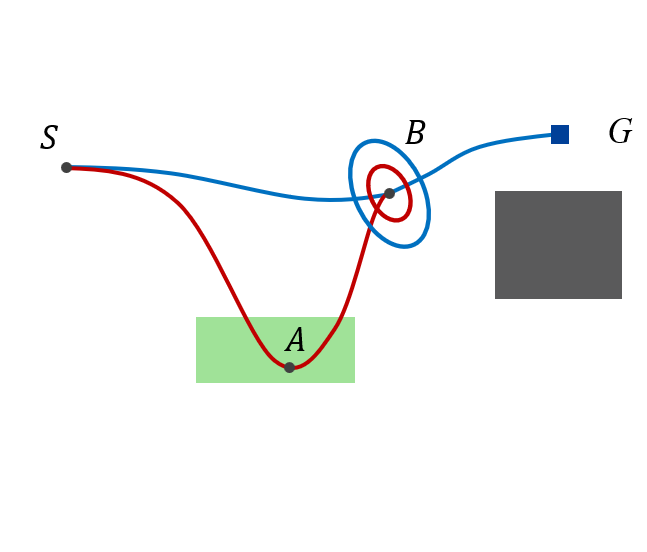}
         \caption{}
     \end{subfigure}
        \caption{(a) RRBT: Two paths reach the same point $B$. The red path detours to an information-rich region to reduce uncertainty. Both paths are explored and preserved in the belief tree in RRBT as it finds all non-dominatd belief nodes.
        (b) IBBT avoids exploring unnecessary belief nodes. If the blue path $\overline{BG}$ satisfies the chance constraint, the whole blue path $\overline{SBG}$ satisfies the chance constraint and has a lower cost than the red path $\overline{SABG}$. The operation of finding more paths reaching $B$ with less uncertainty (but larger cost), including the red one, becomes redundant.}
        \label{BBTmotivation}
\end{figure}

The motivation of IBBT is shown in Figure~\ref{BBTmotivation}.
Two paths reach point $B$ in Figure~\ref{BBTmotivation}(a). 
The red path reaches $B$ with a large cost but with low uncertainty. 
The blue path reaches $B$ with a small cost but with high uncertainty. 
In this case, the blue path cannot dominate the red path, as it will incur a high probability of chance constraint violation for future segments of the path. 
Thus, in RRBT, both paths are preserved in the belief tree. More specifically, RRBT will find all non-dominated belief nodes by exhaustively searching the graph.

However, IBBT avoids exhaustive graph search and hence avoids adding unnecessary belief nodes. In Figure~\ref{BBTmotivation}(b), if the blue path $\overline{BG}$ (starting anywhere inside the blue ellipse) satisfies the chance constraint, the blue path $\overline{SBG}$ will be the solution of the problem since it satisfies the chance constraints and has a lower cost than $\overline{SABG}$.
The operation of searching the current graph to find more paths reaching $B$ with less uncertainty (but a higher cost), including the red one, becomes redundant.

Here, we assume that the cost of the nominal trajectory, $\sum_{k=0}^{N-1} J(\bar{x}_k, \bar{u}_k)$, makes most of the cost in (\ref{eq:obj}).
That is, for the path $\overline{BG}$, starting from the red ellipse and the blue ellipse will incur a similar cost.
Reducing the uncertainty at node $B$ is mainly for satisfying the chance constraint of the future trajectory. 
Such an assumption can also be found, for example, in~\cite{Ichter2017Real}.

RRBT performs an exhaustive search to find all non-dominated nodes whenever a vertex is added to the graph.
Specifically, RRBT will spend a lot of effort finding nodes with low uncertainty but a high cost-to-come.
Such nodes are only necessary if they are indeed part of the optimal path.
If the blue path in Figure~\ref{BBTmotivation}(b) is the solution, we do not need to search for other non-dominated nodes (red ellipse).
However, since we do not know if the future blue path $\overline{BG}$ will satisfy the chance constraint or not, the red node may still be needed.
Thus, IBBT explores the graph and adds belief nodes to the belief tree only when is necessary. This is done by batch sampling and using an informed heuristic.

\subsection{Nominal Trajectory Graph}
The stochastic motion planning problem (\ref{eq:obj})-(\ref{eq:nonlinearModel1}) is divided into a simpler deterministic planning problem and a belief tree search problem. The deterministic planning problem is given by
\begin{align}
& \mathop{\arg\min}_{\bar{u}_k} \ \sum_{k=0}^{N-1} J(\bar{x}_k, \bar{u}_k), \label{DeterministicObj}\\
\mathrm{s.t.} \ \ & \bar{x}_0 = \bar{x}_s, \ \bar{x}_N=\bar{x}_g, \\
& \bar{x}_k \notin \mathcal{X}_\mathrm{obs}, \ k = 0, \cdots, N, \\
& \bar{x}_{k+1} = f(\bar{x}_k,\bar{u}_k,0).
\end{align}

The deterministic planning problem can be solved using sampling-based methods.
The Rapidly-exploring Random Graph (RRG)~\cite{Karaman2011Sampling} algorithm is adopted to add a batch of samples and maintain a graph of nominal trajectories. Similarly, the PRM algorithm \cite{Kavraki1996Prob} may be used in place of RRG. 

%%%%%%%%%%%%%%%%%%%%%%%%%%%%%%%% RRG-D %%%%%%%%%%%%%%%%%%%%%%%%%%%%%%%%%%%%%%%
\IncMargin{.5em}
\begin{algorithm}
\caption{RRG-D$(G,m)$}
\label{alg:RRG-D}

\For{$i=1:m$}
{
    $\bar{x}_\mathrm{rand} \leftarrow \texttt{SampleFree}$\;
    $v_\mathrm{nearest} \leftarrow \texttt{Nearest}(V, \bar{x}_\mathrm{rand})$\;
    $e_\mathrm{nearest} \leftarrow \texttt{Connect}(v_\mathrm{nearest}.\bar{x}, \bar{x}_\mathrm{rand})$\;
    \If{$\texttt{ObstacleFree}(e_\mathrm{nearest})$}
    {
        $V_\mathrm{near} \leftarrow \texttt{Near}(V, \bar{x}_\mathrm{rand})$\;
        $V \leftarrow V \cup \{ v(\bar{x}_\mathrm{rand}) \}$\;
        $E \leftarrow E \cup \{ e_\mathrm{nearest} \}$\;
        $e \leftarrow \texttt{Connect}(\bar{x}_\mathrm{rand}, v_\mathrm{nearest}.\bar{x})$\;
        \If{$\texttt{ObstacleFree}(e)$}
        {
            $E \leftarrow E \cup \{ e \}$\;
        }
        \ForEach{$v_\mathrm{near} \in V_\mathrm{near}$}
        {
            $e \leftarrow \texttt{Connect}(v_\mathrm{near}.\bar{x}, \bar{x}_\mathrm{rand})$\;
            \If{$\texttt{OstaclebFree}(e)$}
            {
                $E \leftarrow E \cup \{ e \}$\;
            }
            $e \leftarrow \texttt{Connect}(\bar{x}_\mathrm{rand}, v_\mathrm{near}.\bar{x})$\;
            \If{$\texttt{ObstacleFree}(e)$}
            {
                $E \leftarrow E \cup \{ e \}$\;
            }
        }
    }
}
\KwRet $G, V_\mathrm{new}$\;
\end{algorithm}
\DecMargin{.5em}
%%%%%%%%%%%%%%%%%%%%%%%%%%%%%%%%%%%%%%%%%%%%%%%%%%%%%%%%%%%%%%%%%%%%%%%%%%%

The RRG-D algorithm given by Algorithm~\ref{alg:RRG-D} follows the RRG algorithm developed in \cite{Karaman2011Sampling} with the additional consideration of system dynamics. 
RRG-D uses the \texttt{Connect} function introduced in Section~\ref{SecProblemformulation} to build a graph of nominal trajectories.
The edge is added to the graph only if the nominal trajectory is obstacle-free, which is indicated by the \texttt{ObstacleFree} checking in Algorithm~\ref{alg:RRG-D}.
RRG-D draws $m$ samples whenever it is called by the IBBT algorithm.  
The $m$ samples constitute one batch. 
The sampled states $\bar{x}$ along with the edges $e$ connecting them generate a graph in the search space. 
For belief space planning, each vertex $v$ has both state information $v.\bar{x}$ and belief information $v.N$. We use $v(\bar{x})$ to refer to the vertex $v$ whose state is $v.\bar{x}$.
RRG-D returns the updated new graph and the newly added vertex set $V_\mathrm{new}$.

\subsection{IBBT}

The Informed Batch Belief Tree algorithm repeatedly performs two main operations: 
It first builds a graph of nominal trajectories to explore the state space of the robot, 
and then it searches over this graph to grow a belief tree in the belief space.
The IBBT algorithm is given by Algorithm~\ref{alg:IBBT} and Algorithm~\ref{alg:GS}.

% For graph construction, batches of samples are added to the graph intermittently. 
% The operation of graph construction is referred to as exploration, as it will incrementally build a graph to cover the state space. Analogously, the operation of searching over the current graph is referred to as exploitation, as it exploits the current graph to search a tree in belief space.
% Previous work \cite{Bry2011Rapidly} performed an exhaustive search whenever one sample is added to the graph, which results in poor performance in terms of exploration. Note that the operation of an exhaustive search is more complicated than adding a single sample to the RRG graph. The proposed Batch Belief Tree (BBT) algorithm adds a batch of samples at a time and it does not require a complete graph search before adding another batch of samples. As it will be shown in Section~\ref{SecExperiment}, the resulting advantage of IBBT is that it finds a better path given the same amount of time compared to \cite{Bry2011Rapidly}. 

%%%%%%%%%%%%%%%%%%%%%%%%% Informed Batch Belief Tree %%%%%%%%%%%%%%%%%%%%%%%%%%%%%%%%%
\IncMargin{.5em}
\begin{algorithm}
\caption{Informed Batch Belief Tree}
\label{alg:IBBT}
$n.P \leftarrow P_0$; $n.\tilde{P} \leftarrow \tilde{P}_0$; $n.c \leftarrow 0$; $n.h \leftarrow \mathrm{Inf}$; $n.\mathrm{parent} \leftarrow \mathrm{null}$\;
$v_s.N \leftarrow \{n\}$; $v_g.N \leftarrow \emptyset$\; 
$v_s.\bar{x} \leftarrow \bar{x}_s$; $v_g.\bar{x} \leftarrow \bar{x}_g$\; 
$v_s.h \leftarrow \mathrm{Inf}$; $v_g.h \leftarrow 0$\; 
$V \leftarrow \{v_s, v_g\}$; $E \leftarrow \emptyset$; $G \leftarrow (V, E)$\;
$Q \leftarrow \{n\}$; $Cost \leftarrow \mathrm{Inf}$\;
\Repeat{$\mathrm{Stop}$}
{
    $(G, V_\mathrm{new}) = \texttt{RRG-D}(G,m)$\;
    $G = \texttt{ValueIteration}(G)$\;
    \ForEach{$v_\mathrm{new} \in V_\mathrm{new}$}
    {
        \ForEach{$v_\mathrm{neighbor} \ \mathrm{of} \ v_\mathrm{new}$}
        {
            $Q \leftarrow Q \cup v_\mathrm{neighbor}.N$\;
        }
    }
    $\texttt{Prune}(Q, Cost)$\;
    $Q \leftarrow Q \cup v_g.N$\;
    $(G, Q, \mathrm{flag}) = \texttt{GraphSearch}(G, Q)$\;
    \If{$\mathrm{flag}$}
    {
        $Cost = \min\{ n.c | \forall n \in v_g.N\}$\;
    }
}
\KwRet $G, \mathrm{flag}$;
\end{algorithm}
\DecMargin{.5em}
%%%%%%%%%%%%%%%%%%%%%%%%%%%%%%%%%%%%%%%%%%%%%%%%%%%%%%%%%%%%%%%%%%%%%%%%%%%%

%%%%%%%%%%%%%%%%%%%%%%%%% Graph Search %%%%%%%%%%%%%%%%%%%%%%%%%%%%%%%%%
\IncMargin{.5em}
\begin{algorithm}
\caption{Graph Search}
\label{alg:GS}
$\mathrm{flag} \leftarrow \mathbf{False}$\;
\While{$Q \neq \emptyset$}
{
    $n \leftarrow \texttt{Pop}(Q)$\;
    \If{$v(n).\bar{x} = \bar{x}_g$}
    {
        $\mathrm{flag} \leftarrow \mathbf{True}$\;
        \KwRet $G, Q, \mathrm{flag}$;
    }
    \ForEach{$v_\mathrm{neighbor} \ \mathrm{of} \ v(n)$}
    {
        $n_\mathrm{new} \leftarrow \texttt{Propagate}(e_\mathrm{neighbor}, n)$\;
        $\mathrm{succ}, G \leftarrow \texttt{AppendBelief}(G, v_\mathrm{neighbor}, n_\mathrm{new})$\;
        \If{$\mathrm{succ}$}
        {
            $Q \leftarrow Q \cup \{n_\mathrm{new}\}$\;
        }
    }
}
\KwRet $G, Q, \mathrm{flag}$;
\end{algorithm}
\DecMargin{.5em}
%%%%%%%%%%%%%%%%%%%%%%%%%%%%%%%%%%%%%%%%%%%%%%%%%%%%%%%%%%%%%%%%%%%%%%%%%%%%

Additional variables are needed to define a belief tree.
A belief node $n$ is defined by a state covariance $n.P$, an estimation error covariance $n.\tilde{P}$, a cost-to-come $n.c$, a heuristic cost-to-go $n.h$, and a parent node index $n.\mathrm{parent}$.
A vertex $v$ is defined by a state $v.\bar{x}$, a set of belief nodes $v.N$, and a vertex cost $v.h$.

The graph search given by Algorithm \ref{alg:GS} repeats two primitive procedures to grow a belief tree: \textit{Belief node selection} which selects the best node in the belief queue for expansion; \textit{Belief propagation} which propagates the selected belief node to its neighbor vertices to generate new belief nodes.
The metric to rank the belief nodes in the belief queue is vital for efficient graph search.

\begin{figure}[htb]
    \centering
    \begin{subfigure}[b]{0.7\columnwidth}
         \centering
         \includegraphics[width=1\columnwidth]{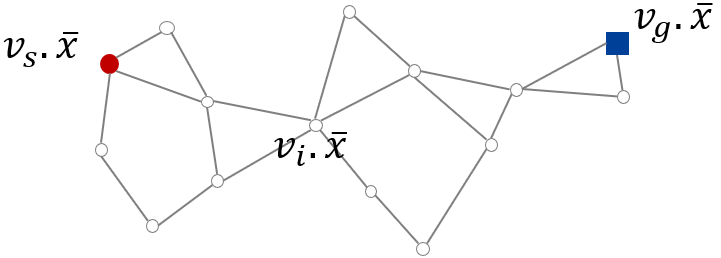}
         \caption{}
     \end{subfigure}
     \begin{subfigure}[b]{0.7\columnwidth}
         \centering
         \includegraphics[width=1\columnwidth]{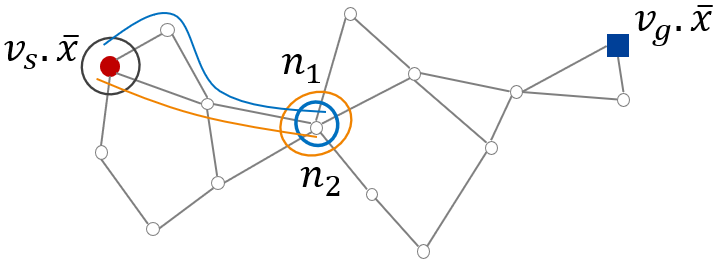}
         \caption{}
     \end{subfigure}
        \caption{(a) Nominal trajectory graph. Each edge is computed by solving a deterministic optimal control problem with edge cost given by (\ref{DeterministicObj}). 
        (b) Two belief nodes are shown at vertex $v_i$.}
        \label{Heuristic}
\end{figure}

Based on the nominal trajectory graph, we can compute the cost-to-go for all vertices. A nominal trajectory graph is shown in Figure \ref{Heuristic}(a). Every edge in the graph is computed by solving a deterministic optimal control problem with edge cost given by (\ref{DeterministicObj}). We compute the cost-to-go $v_i.h$ using value iteration for every vertex in the graph.
$v_i.h$ is the true cost-to-go for the nominal trajectory graph and is an informed, admissible cost-to-go heuristic for the belief tree search problem. 
Here we assume that $J(\bar{x}_k, \bar{u}_k) \leq \mathbb{E} \left[ J(x_k, u_k) \right]$,
thus $\sum_{k=0}^{N-1} J(\bar{x}_k, \bar{u}_k) \leq \mathbb{E}\left[\sum_{k=0}^{N-1} J(x_k, u_k) \right]$. This assumption is true when $J(x_k, u_k)$ is a convex function by using Jensen's inequality \cite{Feller1991An}. For example, a quadratic cost is very common in robotics applications where we want to minimize control effort and state uncertainty (covariance).
Moreover, additional chance constraint checking is performed in the belief tree search. Therefore, $\sum_{k=0}^{N-1} J(\bar{x}_k, \bar{u}_k)$ is an underestimate of the actual cost.  

The nodes in the belief node queue are ranked based on the total heuristic cost $n.f = n.c + n.h$. All belief nodes at the same vertex have the same heuristic cost-to-go and $n.h = v.h$. In Figure \ref{Heuristic}(b), two belief nodes $n_1$, $n_2$ are shown at vertex $v_i$. Their total heuristic costs are $n_1.f = n_1.c + v_i.h$ and $n_2.f = n_2.c + v_i.h$, respectively.

The partial ordering of belief nodes is defined as follows \cite{Bry2011Rapidly}.
Let $n_a$ and $n_b$ be two belief nodes of the same vertex $v$. 
We use $n_a < n_b$ to denote that belief node $n_b$ is dominated by $n_a$. $n_a < n_b$ is true if 
\begin{equation}
    (n_{a}.f < n_{b}.f) \land (n_{a}.P < n_{b}.P) \land (n_{a}.\tilde{P} < n_{b}.\tilde{P}).
\end{equation}
In this case, $n_a$ is better than $n_b$ since it traces back a path that reaches $v$ with less cost and less uncertainty compared with $n_b$.
Next, we summarize some primitive procedures used in the IBBT algorithm. \\
%%%%
\textbf{Pop:} $\texttt{Pop}(Q)$ selects the best belief node in term of the lowest cost $n.f$ from belief queue $Q$ and removes it from $Q$. \\
%%%%
\textbf{Propagate:} The \texttt{Propagate} procedure implements three operations: covariance propagation, chance constraint evaluation, and cost calculation. 
$\texttt{Propagate}(e, n)$ performs the covariance propagation using (\ref{KFupdatea})-(\ref{estimatedstateCov}). 
It takes an edge $e$ and a belief node $n$ at the starting vertex of the edge as inputs.
Chance constraints are evaluated using the state covariance $P_k$ along the edge.
If there are no chance constraint violations, a new belief $n_\mathrm{new}$ is returned, which is the final belief at the end vertex of the edge.
Otherwise, the procedure returns no belief.
The cost-to-come of $n_\mathrm{new}$ is the sum of $n.c$ and the cost of edge $e$ by applying the controller (\ref{eq:feedbackControl}) associated with $e$. \\
%%%%
\textbf{Append Belief:} The function \texttt{AppendBelief}$(G,v,n_\mathrm{new})$ decides if the new belief $n_\mathrm{new}$ should be added to vertex $v$ or not.
If $n_\mathrm{new}$ is not dominated by any existing belief nodes in $v.N$, $n_\mathrm{new}$ is added to $v.N$.
Note that adding $n_\mathrm{new}$ means extending the current belief tree such that $n_\mathrm{new}$ becomes a leaf node of the current belief tree.
Next, we also check if any existing belief node in $v.N$ is dominated by $n_\mathrm{new}$.
If an existing belief is dominated, its descendant and the node itself are pruned. \\
\textbf{Prune Node Queue:} The function \texttt{Prune}$(Q,Cost)$ removes nodes in $Q$ whose total heuristic cost is greater than $Cost$. $Cost$ is the cost of the current solution found.\\
\textbf{Value Iteration:} The function \texttt{ValueIteration}$(G)$ computes the cost-to-go for all vertices in $G$ use value iteration. The value iteration is done using the nominal trajectory graph. For vertices whose cost-to-go values are computed in the last iteration (before calling this function), their values are reused for initialization for faster convergence. \\

In Algorithm \ref{alg:IBBT}, Line 1-5 initializes the graph and the belief tree.
The initial condition of the motion planning problem is given by the starting state $\bar{x}_s$, state covariance $P_0$, and estimation error covariance $\tilde{P}_0$. The goal state is $\bar{x}_g$. In Line 6, the queue $Q$ is initialized with the initial node $n$ and the cost of the current solution is set as infinity.
In Line 8, the RRG-D is called to add $m$ samples and maintain a graph of nominal trajectories, $V_\mathrm{new}$ is the set of newly added vertices after calling RRG-D.
Based the on the nominal trajectory graph, cost-to-go for all vertices in $G$ is computed using value iteration (Line 9). 
Line 10-12 update the belief node queue after batch sampling. For every vertex that has an outgoing edge towards $v_\mathrm{new}$, all the belief nodes at that vertex are added to the queue.

In Algorithm \ref{alg:GS}, the belief $n$ is propagated outwards to all the neighbor vertices of $v(n)$ to grow the belief tree in Line 7-11. $v(n)$ refers to the vertex associated with $n$.
$v_\mathrm{neighbor}$ is a neighbor of $v(n)$ when there is an edge $e_\mathrm{neighbor}$ from $v(n)$ to $v_\mathrm{neighbor}$ in the graph.
The new belief $n_\mathrm{new}$ is added to the $v_{\mathrm{neighbor}}.N$ and $Q$ if the belief tree extension is successful. Then, $n$ is marked as the parent node of $n_\mathrm{new}$.
Note that each belief node traces back a unique path from the initial belief node. For every belief node in the belief tree, we already found a feasible path (satisfies chance constraint) to this node. 
Algorithm \ref{alg:GS} terminates when the belief node at $\bar{x}_g$ is selected for expansion (Line 4-6, Algorithm \ref{alg:GS}) or $Q$ is empty. In the first case, the best solution is found. In the second case, no solution exists given the current graph.

\begin{comment}
\subsection{Convergence Analysis} \label{SubSecAnalysis}

In this section, we briefly discuss the asymptotic optimality of the IBBT algorithm, which states that the solution returned by IBBT converges to the optimal solution as the number of the samples goes to infinite.

Note that the RRBT algorithm is shown to be asymptotically optimal~\cite{Bry2011Rapidly}. 
Here, we argue the asymptotic optimality of the IBBT by drawing to its connection with the RRBT algorithm.

\begin{proposition} \label{Prop1}
Given the same sequence of samples of the RRBT algorithm with any fixed length, the RRG graphs constructed by IBBT and RRBT are the same.
Provided that the terminate condition in Line 13 of Algorithm~\ref{alg:IBBT} is not satisfied, the IBBT algorithm finds all the non-dominated belief nodes over the RRG graph and the belief trees built by IBBT and RRBT are the same.  
% \begin{proof}
% See Appendix \ref{AppendixForProp1}.
% \end{proof}
\end{proposition}

\begin{proof}
The proof of Proposition~\ref{Prop1} is straightforward.
After running the IBBT algorithm, both $Q_\mathrm{current}$ and $Q_\mathrm{next}$ will be empty.
The \texttt{AppendBelief} function ensures that every  non-dominated belief is added to the belief tree, and only the dominated beliefs are pruned.
Thus, the RRG graph is exhaustively searched and all the non-dominated belief nodes are in the belief tree.
Therefore, the asymptotic optimality of the RRBT algorithm implies the asymptotic optimality of the IBBT algorithm.
\end{proof}
\end{comment}

\section{Experimental Results} \label{SecExperiment}

In this section, we test the IBBT algorithm for different motion planning problems and compared the results with the RRBT algorithm~\cite{Bry2011Rapidly}.
\begin{figure}[htb]
    \centering
    \begin{subfigure}[b]{0.38\columnwidth}
         \centering
         \includegraphics[width=1\columnwidth]{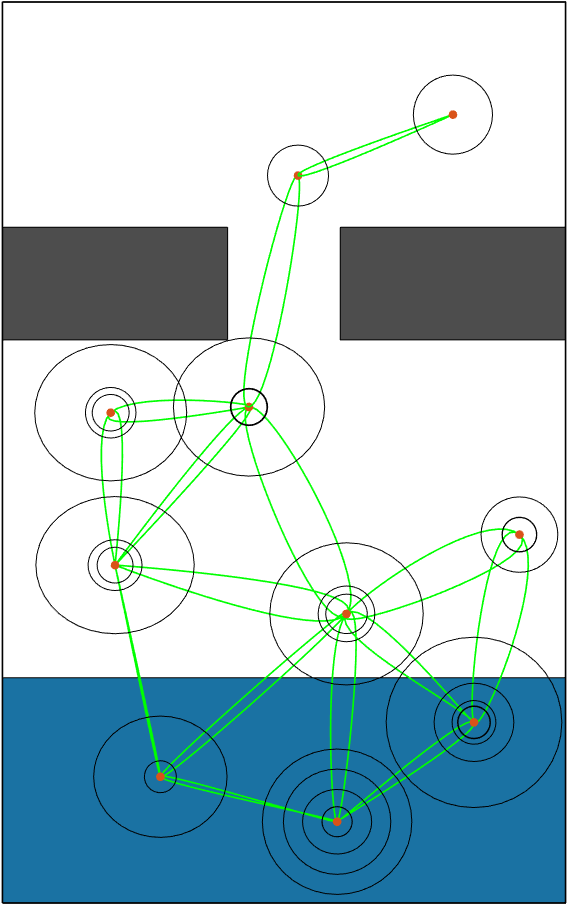}
         \caption{}
     \end{subfigure}
     \begin{subfigure}[b]{0.38\columnwidth}
         \centering
         \includegraphics[width=1\columnwidth]{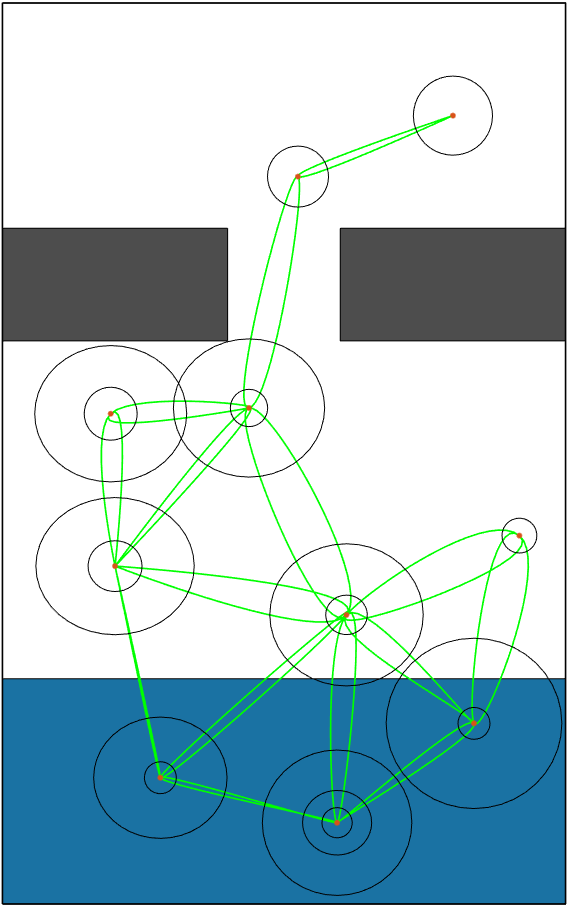}
         \caption{}
     \end{subfigure}
        \caption{(a) Nominal trajectory graph and belief tree from the RRBT algorithm. (b) Nominal trajectory graph and belief tree from the IBBT algorithm. Both algorithms stop when they find the first solution. 
        The extra ellipses in the left figure indicate that RRBT adds more nodes to the belief tree by exhaustive search. 
        IBBT avoids unnecessary belief nodes expansion and find the same solution faster. }
        \label{FirstEnv:tree}
\end{figure}

\begin{figure}[htb]
    \centering
    \includegraphics[width=0.38\columnwidth]{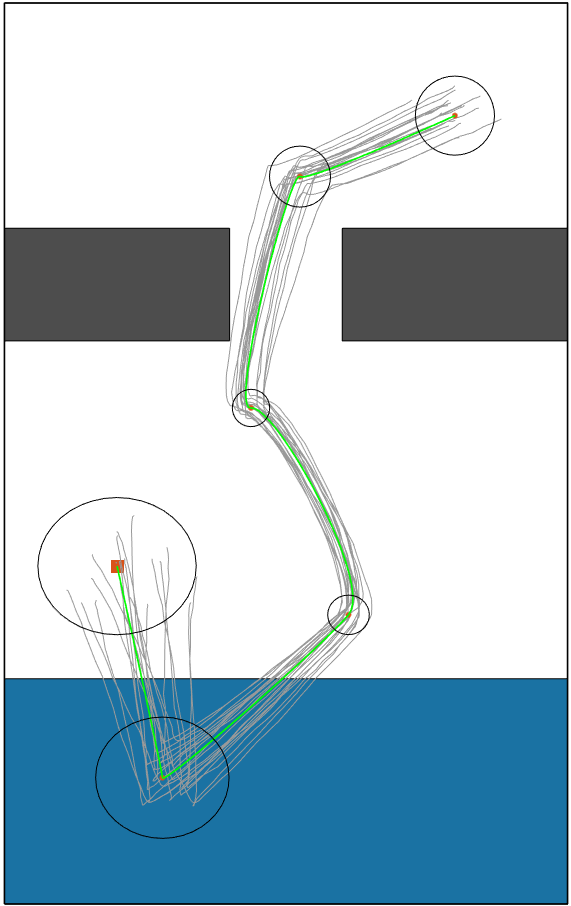}
    \caption{First solution found by both algorithms.}
    \label{FirstEnv:result}
\end{figure}

\begin{figure}[htb]
    \centering
    \includegraphics[width=0.6\columnwidth]{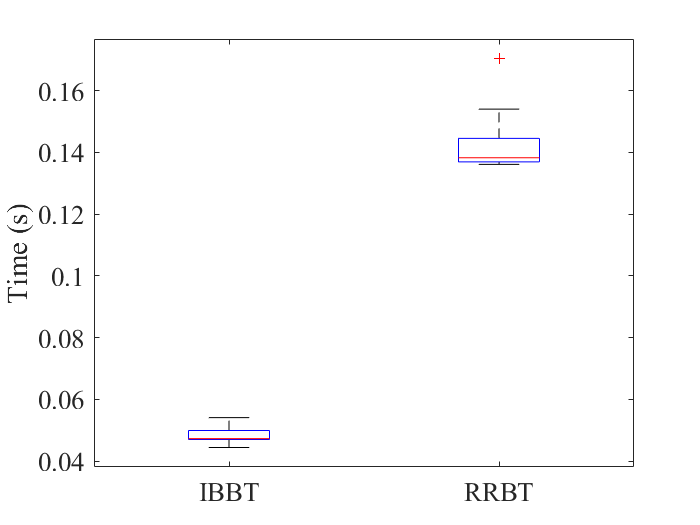}
    \caption{Comparison between the IBBT and the RRBT algorithms. IBBT is faster to find the first solution.}
    \label{FirstEnv:compare}
\end{figure}

\subsection{Double Integrator}

The first planning environment is shown in Figure~\ref{FirstEnv:tree}.  
The gray areas are obstacles and
the blue region is the information-rich region, that is,
the measurement noise is small when the robot is in this region.
We use the 2D double integrator dynamics with motion and sensing uncertainties as an example.
The system model is linear and is given by
\begin{equation}
\begin{split}
    {x}_{k+1} &= A_k x_k + B_k u_k + G_k w_k, \\
    y_k &= C_k x_k + D_k v_k,
    \label{DoubleIntegrator}
\end{split}
\end{equation}
where the system state includes position and velocity, the control input is the acceleration.
The system matrices are given by 
\begin{equation}
    A_k = \begin{bmatrix} 1 & 0 & \Delta t & 0 \\ 0 & 1 & 0 & \Delta t \\ 0 & 0 & 1 & 0 \\ 0 & 0 & 0 & 1 \end{bmatrix}, \quad
    B_k = \begin{bmatrix} \Delta t^2/2 & 0 \\ 0 & \Delta t^2/2 \\ \Delta t & 0 \\ 0 & \Delta t \end{bmatrix}, \quad
    C_k = I_4,
\end{equation}
$G_k = \sqrt{\Delta t} \, \mathrm{diag}(0.03, 0.03, 0.02, 0.02)$, and $D_k = 0.01 I_4$ when the robot is in a information-rich region, otherwise $D_k = I_4$.
%$I_n$ is an identity matrix with dimension $n$.

%To compute the nominal trajectory, we consider a quadratic cost of the control input where the cost matrix is $R = I_2$.
To compute the nominal trajectories, the analytical solution is available~\cite{Zheng2021Belief}.
An LQG controller is used to compute the feedback gain $K$ in the \texttt{Connect} function.
The collision probability in the chance constraint is approximated using Monte Carlo simulations.
We sample from the state distribution and count the number of samples that collide with the obstacles. 
The ratio of collided samples to the total samples is the approximate collision probability. 
%All simulations were done on a laptop computer with a 1.6 GHz Intel i5-8255u processor and 8 GB RAM using MATLAB.

We compared the performance of RRBT and IBBT to find the first solution.
The belief tree from RRBT is shown in Figure~\ref{FirstEnv:tree}(a), and the belief tree from IBBT is shown in Figure~\ref{FirstEnv:tree}(b).
Both algorithms use the same set of states and find the same solution, which is given in Figure~\ref{FirstEnv:result}.
The robot first goes down to the information-rich region to reduce its uncertainty, while directly moving toward to goal will violate the chance constraint.

Fewer belief nodes are searched and added to the tree using IBBT compared with RRBT, even though they return the same solution.
IBBT uses batch sampling and computes the informed cost-to-go heuristic to guide the belief tree search, while RRBT only uses the cost-to-come.
RRBT tries to find all non-dominated belief nodes whenever a vertex is added to the graph. Thus, it will find belief nodes that have low uncertainty but high cost-to-come (shown as small ellipses in Figure~\ref{FirstEnv:tree}(a)). 
However, if such a node is not part of the solution path, this computation is not necessary. 
The comparison of the results is shown in Figure~\ref{FirstEnv:compare}.
The solving time for IBBT and RRBT is around 0.05 sec and 0.14 sec respectively.

\begin{figure}[htb]
    \centering
    \begin{subfigure}[b]{0.58\columnwidth}
         \centering
         \includegraphics[width=\columnwidth]{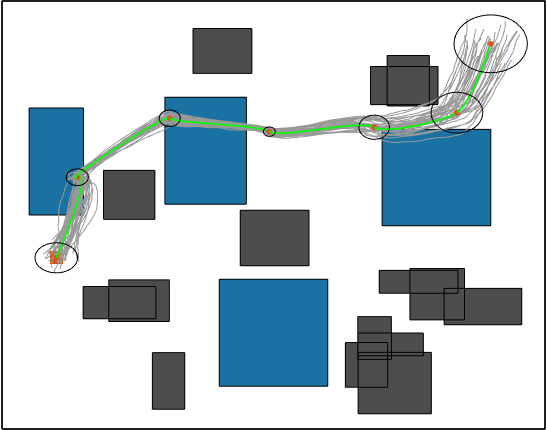}
         \caption{}
     \end{subfigure}
         \centering
    \begin{subfigure}[b]{0.58\columnwidth}
         \centering
         \includegraphics[width=\columnwidth]{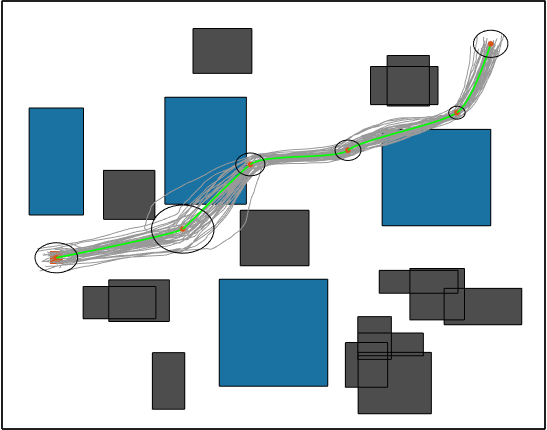}
         \caption{}
     \end{subfigure}
        \caption{Planning results of double integrator. (a) The first solution returned by IBBT; (b) Final solution with solving time less than 2 sec.}
        \label{SecondEnv}
\end{figure}

\begin{figure}[htb]
    \centering
    \includegraphics[width=0.66\columnwidth]{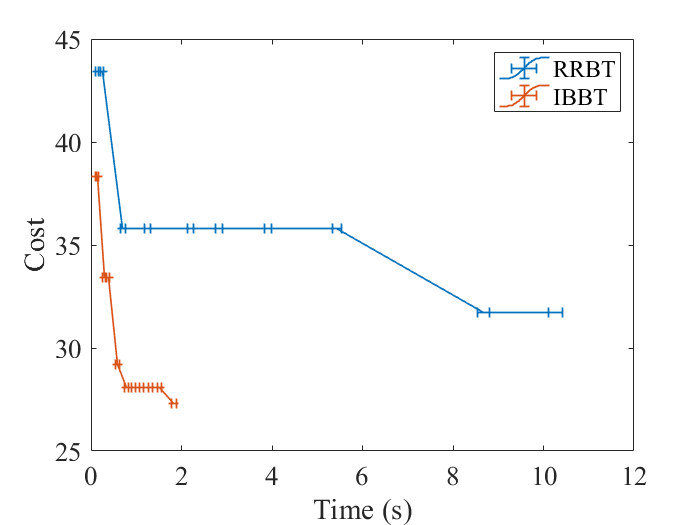}
    \caption{Comparison between IBBT and RRBT. IBBT has better cost-time performance and finds the first solution with less time.}
    \label{SecondEnv:compare}
\end{figure}

The second planning environment is shown in Figure~\ref{SecondEnv}.
The problem setting is similar to the first environment except that more obstacles and information-rich regions are added.
The first solution and the improved solution are shown in Figure~\ref{SecondEnv}(a) and Figure~\ref{SecondEnv}(b), respectively.
The green lines are the mean trajectories.
The gray lines around the green lines are the Monte-Carlo simulation results.
The comparison with the RRBT algorithm is given in Figure~\ref{SecondEnv:compare}. 
After finding the initial solution, both algorithms are able to improve their solution when more samples are added to the graph but
IBBT is able to find a better solution in a much shorter time.

\subsection{Dubins Vehicle}

Finally, we tested our algorithm using the Dubins vehicle model.
The deterministic discrete-time model is given by
\begin{equation}
\begin{split}
    {x}_{k+1} &= x_k + \cos{\theta_{k}} \Delta t, \\
    {y}_{k+1} &= y_k + \sin{\theta_{k}} \Delta t, \\
    {\theta}_{k+1} &= \theta_k + u_k \Delta t.
    \label{Eq:Dubin}
\end{split}
\end{equation}
The nominal trajectory for the Dubins vehicle is chosen as the minimum length path connecting two configurations of the vehicle.
The analytical solution for the nominal trajectory is available in \cite{LaValle2006Planning}.

After linearization, the error dynamics around the nominal path is given by (\ref{DoubleIntegrator}), where the system matrices are
\begin{equation}
    A_k = \begin{bmatrix} 1 & 0 & -\sin{\theta_k} \Delta t \\ 0 & 1 & \cos{\theta_k}\Delta t \\ 0 & 0 & 1 \end{bmatrix}, \quad
    B_k = \begin{bmatrix} 0 \\ 0 \\ \Delta t \end{bmatrix}, \quad
    C_k = I_3. 
\end{equation}
$G_k = \sqrt{\Delta t}\, \mathrm{diag}(0.02, 0.02, 0.02)$, $D_k = 0.1 I_3$ when the robot is in a information-rich region, otherwise $D_k = 2 I_3$.
An LQG controller is used to compute the feedback gain $K$, the weighting matrices of the LQG cost are $Q = 2 I_3$ and $R = 1$.

\begin{figure}[htb]
    \centering
    \begin{subfigure}[b]{0.58\columnwidth}
         \centering
         \includegraphics[width=\columnwidth]{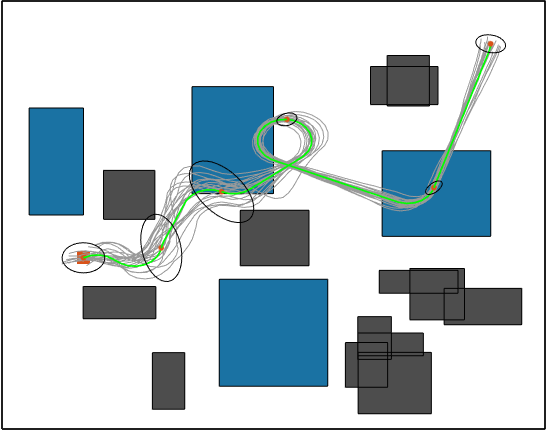}
         \caption{}
     \end{subfigure}
         \centering
    \begin{subfigure}[b]{0.58\columnwidth}
         \centering
         \includegraphics[width=\columnwidth]{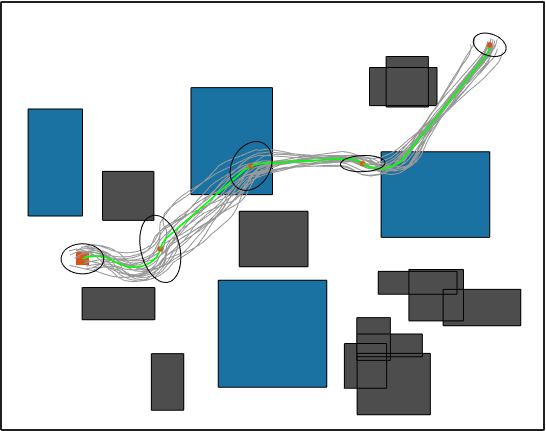}
         \caption{}
     \end{subfigure}
        \caption{Planning results of the Dubins vehicle. (a) First solution. (b) Final solution.
}
        \label{Fig:Dubin}
\end{figure}

\begin{figure}[htb]
    \centering
    \includegraphics[width=0.66\columnwidth]{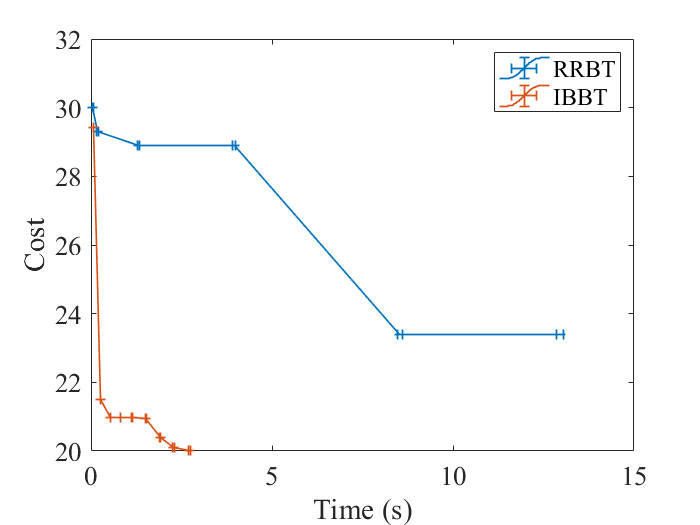}
    \caption{Comparison between IBBT and RRBT. IBBT has better cost-time performance and finds the first solution with less time.}
    \label{Dubin:compare}
\end{figure}

The first solution and the improved solution are shown in Figure~\ref{Fig:Dubin}(a) and Figure~\ref{Fig:Dubin}(b), respectively.
The green line is the mean trajectory.
The gray lines around the green lines are the Monte-Carlo simulations.
The comparison with the RRBT algorithm is given in Figure~\ref{Dubin:compare}. 
After finding the initial solution, both algorithms are able to improve their current solution when more samples are added to the graph.
Again, IBBT has better cost vs. time performance.

\section{Conclusion}  \label{secConclusion}
We developed an online, anytime, incremental algorithm, IBBT, for motion planning under uncertainties. The algorithm considers a robot that is partially observable, has motion uncertainty, and operates in a continuous domain.
The algorithm interleaves between batch sampling, building a graph of nominal trajectories in the state space, and searches over the graph to grow a belief tree. 
The heuristic cost-to-go is computed using the nominal trajectory graph along with value iteration. This cost-to-go along with the cost-to-come provides an informed heuristic to guide the belief tree search.
The algorithm finds motion plans that converge to the optimal one as more batches of samples are added to the graph.
We have tested the IBBT algorithm in different planning environments. 
The proposed algorithm finds non-trivial motion plans and provides better solutions using a smaller amount of time compared with previous methods.
%\section*{Acknowledgments}

\bibliographystyle{ieeetr}
\bibliography{references}

\begin{thebibliography}{10}

\bibitem{Thrun2005Probabilistic}
S.~Thrun, W.~Burgard, and D.~Fox, {\em Probabilistic Robotics}.
\newblock MIT Press, 2005.

\bibitem{Van2012Motion}
J.~Van Den~Berg, S.~Patil, and R.~Alterovitz, ``Motion planning under
  uncertainty using iterative local optimization in belief space,'' {\em The
  International Journal of Robotics Research}, vol.~31, no.~11, pp.~1263--1278,
  2012.

\bibitem{Kaelbling1998Planning}
L.~P. Kaelbling, M.~L. Littman, and A.~R. Cassandra, ``Planning and acting in
  partially observable stochastic domains,'' {\em Artificial Intelligence},
  vol.~101, pp.~99--134, 1998.

\bibitem{Porta2006Point}
J.~M. Porta, N.~Vlassis, M.~T. Spaan, and P.~Poupart, ``Point-based value
  iteration for continuous {POMDPs},'' {\em Journal of Machine Learning
  Research}, pp.~2329--2367, November 2006.

\bibitem{shani2013survey}
G.~Shani, J.~Pineau, and R.~Kaplow, ``A survey of point-based {POMDP}
  solvers,'' {\em Autonomous Agents and Multi-Agent Systems}, vol.~27, no.~1,
  pp.~1--51, 2013.

\bibitem{somani2013despot}
A.~Somani, N.~Ye, D.~Hsu, and W.~S. Lee, ``{DESPOT:} online {POMDP} planning
  with regularization,'' {\em Advances in neural information processing
  systems}, vol.~26, pp.~1772--1780, 2013.

\bibitem{Karaman2011Sampling}
S.~Karaman and E.~Frazzoli, ``Sampling-based algorithms for optimal motion
  planning,'' {\em The International Journal of Robotics Research}, vol.~30,
  pp.~846--894, June 2011.

\bibitem{Prentice2009The}
S.~Prentice and N.~Roy, ``The belief roadmap: Efficient planning in belief
  space by factoring the covariance,'' {\em The International Journal of
  Robotics Research}, vol.~28, pp.~1448--1465, 2009.

\bibitem{Bry2011Rapidly}
A.~Bry and N.~Roy, ``Rapidly-exploring random belief trees for motion planning
  under uncertainty,'' in {\em IEEE International Conference on Robotics and
  Automation}, pp.~723--730, May 2011.

\bibitem{Luders2013Robust}
B.~D. Luders, S.~Karaman, and J.~P. How, ``Robust sampling-based motion
  planning with asymptotic optimality guarantees,'' in {\em AIAA Guidance,
  Navigation, and Control}, p.~5097, 2013.

\bibitem{Sun2015High}
W.~Sun, S.~Patil, and R.~Alterovitz, ``High-frequency replanning under
  uncertainty using parallel sampling-based motion planning,'' {\em IEEE
  Transactions on Robotics}, vol.~31, no.~1, pp.~104--116, 2015.

\bibitem{Janson2018Monte}
L.~Janson, E.~Schmerling, and M.~Pavone, ``{Monte Carlo} motion planning for
  robot trajectory optimization under uncertainty,'' in {\em Robotics
  Research}, pp.~343--361, Springer, Cham., 2018.

\bibitem{Ichter2017Real}
B.~Ichter, A.~A. Schmerling, E.and Agha-mohammadi, and M.~Pavone, ``Real-time
  stochastic kinodynamic motion planning via multiobjective search on {GPUs},''
  in {\em IEEE International Conference on Robotics and Automation},
  pp.~5019--5026, May 2017.

\bibitem{Agha-Mohammadi2014FIRM}
A.~A. Agha-Mohammadi, S.~Chakravorty, and N.~M. Amato, ``{FIRM:} sampling-based
  feedback motion-planning under motion uncertainty and imperfect
  measurements,'' {\em The International Journal of Robotics Research},
  vol.~33, no.~2, pp.~268--304, 2014.

\bibitem{Blackmore2011Chance}
L.~Blackmore, M.~Ono, and B.~C. Williams, ``Chance-constrained optimal path
  planning with obstacles,'' {\em IEEE Transactions on Robotics}, vol.~27,
  no.~6, pp.~1080--1094, 2011.

\bibitem{Vitus2011Closed}
M.~P. Vitus and C.~J. Tomlin, ``Closed-loop belief space planning for linear,
  {Gaussian} systems,'' in {\em IEEE International Conference on Robotics and
  Automation}, pp.~2152--2159, May 2011.

\bibitem{Wang2020Non-Gaussian}
A.~Wang, A.~Jasour, and B.~C. Williams, ``{Non-Gaussian} chance-constrained
  trajectory planning for autonomous vehicles under agent uncertainty,'' {\em
  IEEE Robotics and Automation Letters}, vol.~5, no.~4, pp.~6041--6048, 2020.

\bibitem{Aoude2013Prob}
G.~S. Aoude, B.~D. Luders, J.~M. Joseph, N.~Roy, and J.~P. How,
  ``Probabilistically safe motion planning to avoid dynamic obstacles with
  uncertain motion patterns,'' {\em Autonomous Robots}, vol.~35, no.~1,
  pp.~51--76, 2013.

\bibitem{Summers2018Distributionally}
T.~Summers, ``Distributionally robust sampling-based motion planning under
  uncertainty,'' in {\em IEEE/RSJ International Conference on Intelligent
  Robots and Systems}, pp.~6518--6523, October 2018.

\bibitem{Zheng2021Belief}
D.~Zheng, J.~Ridderhof, P.~Tsiotras, and A.~A. Agha-mohammadi, ``Belief space
  planning: A covariance steering approach,'' in {\em International Conference
  on Robotics and Automation}, (Philadelphia, PA), pp.~11051--11057, 2022.

\bibitem{janson2015fast}
L.~Janson, E.~Schmerling, A.~Clark, and M.~Pavone, ``Fast marching tree: A fast
  marching sampling-based method for optimal motion planning in many
  dimensions,'' {\em The International Journal of Robotics Research}, vol.~34,
  no.~7, pp.~883--921, 2015.

\bibitem{gammell2020batch}
J.~D. Gammell, T.~D. Barfoot, and S.~S. Srinivasa, ``{Batch Informed Trees
  (BIT*)}: Informed asymptotically optimal anytime search,'' {\em The
  International Journal of Robotics Research}, vol.~39, no.~5, pp.~543--567,
  2020.

\bibitem{Van2011LQG-MP}
J.~Van Den~Berg, P.~Abbeel, and K.~Goldberg, ``{LQG-MP:} optimized path
  planning for robots with motion uncertainty and imperfect state
  information,'' {\em The International Journal of Robotics Research}, vol.~30,
  no.~7, pp.~895--913, 2011.

\bibitem{Platt2010Belief}
R.~Platt, R.~Tedrake, L.~Kaelbling, and T.~Lozano-Perez, ``Belief space
  planning assuming maximum likelihood observations,'' in {\em Robotics:
  Science and Systems}, 2010.

\bibitem{lavalle2001randomized}
S.~LaValle and J.~J. Kuffner~Jr, ``Randomized kinodynamic planning,'' {\em The
  International Journal of Robotics Research}, vol.~20, no.~5, pp.~378--400,
  2001.

\bibitem{Liu2014Incremental}
W.~Liu and M.~H. Ang, ``Incremental sampling-based algorithm for risk-aware
  planning under motion uncertainty,'' in {\em IEEE International Conference on
  Robotics and Automation}, pp.~2051--2058, May 2014.

\bibitem{Luders2010Chance}
B.~Luders, M.~Kothari, and J.~How, ``Chance constrained {RRT} for probabilistic
  robustness to environmental uncertainty,'' in {\em AIAA Guidance, Navigation,
  and Control}, p.~8160, 2010.

\bibitem{Shan2017Belief}
T.~Shan and B.~Englot, ``Belief roadmap search: Advances in optimal and
  efficient planning under uncertainty,'' in {\em IEEE/RSJ International
  Conference on Intelligent Robots and Systems}, pp.~5318--5325, September
  2017.

\bibitem{Safaoui2021Risk}
S.~Safaoui, B.~J. Gravell, V.~Renganathan, and T.~H. Summers, ``Risk-averse
  {RRT*} planning with nonlinear steering and tracking controllers for
  nonlinear robotic systems under uncertainty,'' {\em arXiv preprint
  arXiv:2103.05572}, 2021.

\bibitem{Wang2020Moment}
A.~Wang, A.~Jasour, and B.~Williams, ``Moment state dynamical systems for
  nonlinear chance-constrained motion planning,'' {\em arXiv preprint
  arXiv:2003.10379}, 2020.

\bibitem{Ho2022Gaussian}
Q.~H. Ho, Z.~N. Sunberg, and M.~Lahijanian, ``Gaussian belief trees for chance
  constrained asymptotically optimal motion planning,'' in {\em International
  Conference on Robotics and Automation}, (Philadelphia, PA), pp.~11029--11035,
  2022.

\bibitem{Sun2021Belief}
K.~Sun and V.~Kumar, ``Belief space planning for mobile robots with range
  sensors using {iLQG},'' {\em IEEE Robotics and Automation Letters}, vol.~6,
  no.~2, pp.~1902--1909, 2021.

\bibitem{Rahman2021Uncertainty}
S.~Rahman and S.~L. Waslander, ``Uncertainty-constrained differential dynamic
  programming in belief space for vision based robots,'' {\em IEEE Robotics and
  Automation Letters}, vol.~6, no.~2, pp.~3112--3119, 2021.

\bibitem{Kavraki1996Prob}
L.~E. Kavraki, P.~Svestka, J.~C. Latombe, and M.~H. Overmars, ``Probabilistic
  roadmaps for path planning in high-dimensional configuration spaces,'' {\em
  IEEE Transactions on Robotics and Automation}, vol.~12, no.~4, pp.~566--580,
  1996.

\bibitem{Feller1991An}
W.~Feller, {\em An introduction to probability theory and its applications},
  vol.~2.
\newblock John Wiley and Sons, 1991.

\bibitem{LaValle2006Planning}
S.~M. LaValle, {\em Planning Algorithms}.
\newblock Cambridge university press, 2006.

\end{thebibliography}

\addtolength{\textheight}{-12cm}   % This command serves to balance the column lengths

\end{document}